\documentclass[preprint,12pt]{elsarticle}

\usepackage{amssymb}
\usepackage{amsfonts,amsmath,amscd}
\usepackage{float}
\usepackage{graphicx}
\usepackage{subfig}
\usepackage{nccmath}
\usepackage{siunitx}
\usepackage{booktabs}
\usepackage{url}
\usepackage{caption}

\usepackage{xcolor}

\journal{Simulation Modelling Practice and Theory}

\begin{document}

\begin{frontmatter}



\title{Autonomous Locomotion Mode Transition in Quadruped Track-Legged Robots: A Simulation-Based Analysis for Step Negotiation}


\author[label1]{Jie\ Wang\corref{cor1}}
\ead{jwangjie@outlook.com}
\cortext[cor1]{Corresponding author}
\author[label2]{Krispin\ Davies}
\ead{kdavies@clearpath.ai}

\affiliation[label1]{organization={Electrical and Computer Engineering Department, University of Waterloo},
            addressline={200 University Avenue West}, 
            city={Waterloo},
            postcode={N2L~3G1}, 
            state={ON},
            country={Canada}}

\affiliation[label2]{organization={Clearpath Robotics Inc.},
            addressline={1425 Strasburg Rd. Suite 2A}, 
            city={Kitchener},
            postcode={N2R~1H2}, 
            state={ON},
            country={Canada}}

            
\begin{abstract}
Hybrid track/wheel-legged robots combine the advantages of wheel-based and leg-based locomotion, granting adaptability across varied terrains through efficient transitions between rolling and walking modes. However, automating these transitions remains a significant challenge. In this paper, we introduce a method designed for autonomous mode transition in a quadruped hybrid robot with a track/wheel-legged configuration, especially during step negotiation. Our approach hinges on a decision-making mechanism that evaluates the energy efficiency of both locomotion modes using a proposed energy-based criterion. To guarantee a smooth negotiation of steps, we incorporate two climbing gaits designated for the assessment of energy usage in walking locomotion. Simulation results validate the method's effectiveness, showing successful autonomous transitions across steps of diverse heights. Our suggested approach has universal applicability and can be modified to suit other hybrid robots of similar mechanical configuration, provided their locomotion energy performance is studied beforehand.
\end{abstract}




\end{frontmatter}


\section{Introduction}
\label{section:Introduction}

\begin{figure}
    \vspace{0.3cm}
    \centerline{\includegraphics[width=0.8\columnwidth]{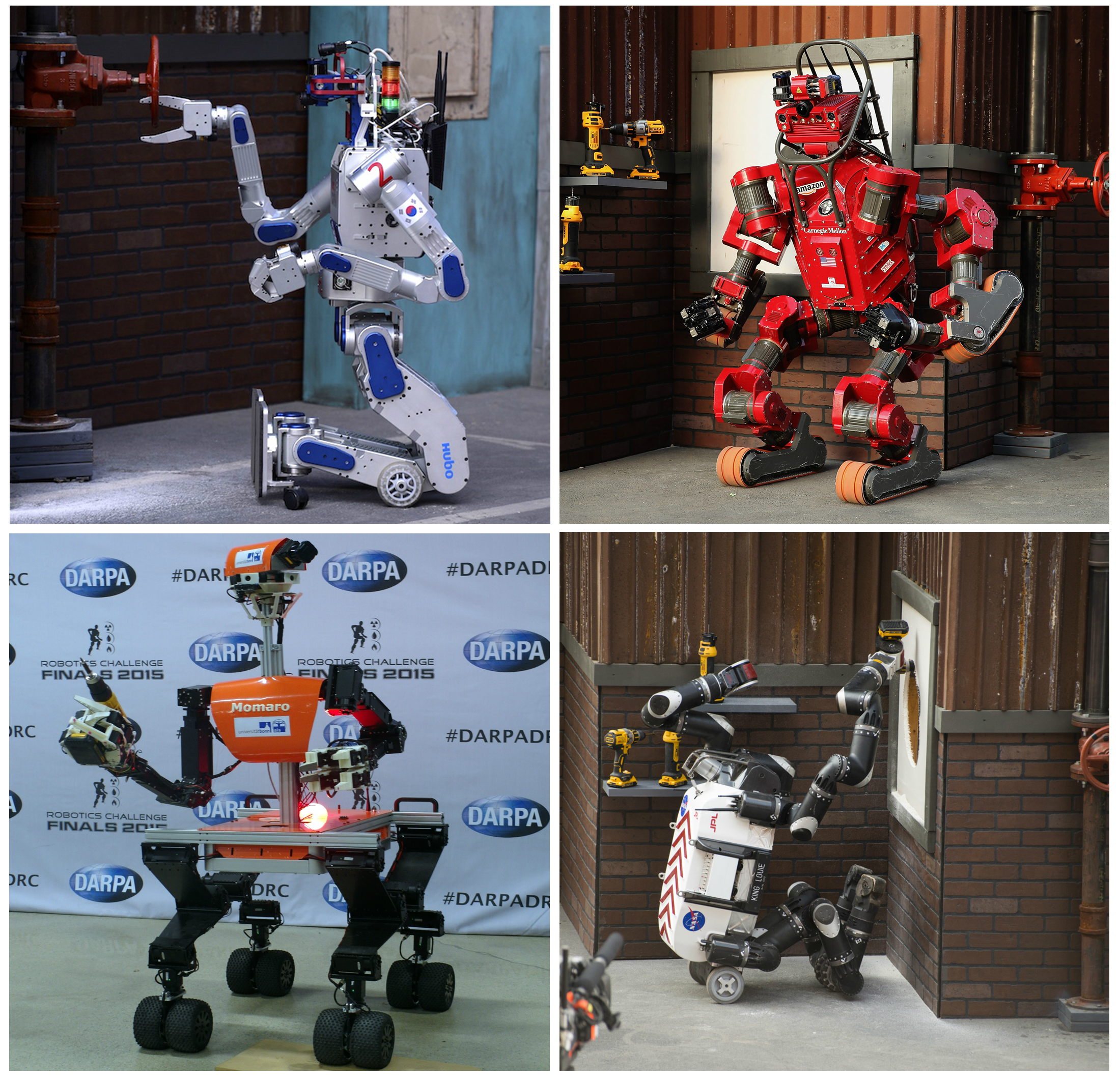}}
    \caption{Four cutting-edge wheel/track-legged robots: from left to right are DRC-HUBO, CHIMP, Momaro, and RoboSimian \cite{krotkov2018darpa}.
    }
    \label{figure:1-1}
\end{figure}

In the realm of mobile robotics research, the motion control of terrestrial robots across varied terrains is a complex endeavor. To enhance locomotion efficacy and elevate mobility, hybrid robots have been actively developed in the past decade \cite{krotkov2018darpa}. These robots astutely choose the most suitable locomotion mode from a spectrum of options based on the terrain \cite{kobayashi2012locomotion}. Hybrid systems that integrate wheels or tracks and legs have been dominant designs due to their unparalleled efficiency and adeptness across diverse landscapes \cite{bjelonic2022survey}. These wheel/track-legged robots, see Fig. \ref{figure:1-1}, seamlessly blend the rolling efficiency of wheels or tracks with the versatile adaptability of legs. On even terrains, they predominantly rely on their wheeling/tracking mechanisms, capitalizing on the energy-saving benefits of rolling motion. However, when these robots encounter rough terrains such as steps and stairs, they have the capability to transition to their legged mode. Although legged locomotion might consume more energy than wheeled travel on flat surfaces, its versatility ensures the robot can traverse challenging areas where purely wheeled mechanisms would falter \cite{wang2017autonomous}. 

There are two primary technical challenges in the wheel/track-legged robotics area \cite{kobayashi2012locomotion}. First, there's a need to ensure accurate motion control within both rolling and walking locomotion modes \cite{okada2010motion} and effectively handle the transitions between them \cite{qiu2022upright}. Second, it's essential to develop decision-making frameworks that determine the best mode---either rolling or walking---based on the robot's environmental interactions and internal states \cite{kobayashi2013locomotion, leppanen2008sensing}. In addressing the first challenge, the dynamics of rolling locomotion are well understood and are similar to those of traditional wheeled/tracked robots. However, despite extensive research on the walking dynamics of standard legged robots, focused studies on the walking patterns specific to wheel/track-legged robots are limited \cite{thomson2012kinematic}. Transition control between these locomotion modes for wheel/track-legged robots also requires more exploration \cite{qiu2022upright}. In this study, we focus on the second challenge to develop efficient decision-making algorithms for transitioning between locomotion modes. This remains a very less explored area \cite{bjelonic2022survey}, but is essential to achieve an autonomous locomotion transition in hybrid robots. Building upon our prior work, we employ two climbing gaits to ensure smooth walking locomotion for wheel/track-legged robots, particularly when navigating steps \cite{wang2016stair}.

Hybrid robots typically transition between locomotion modes either by ``supervised autonomy'' \cite{stentz2015chimp}, where human operators make the switch decisions, or the autonomous locomotion mode transition approach, where robots autonomously swap the modes predicated on pre-set criteria \cite{leppanen2008sensing}. However, the execution of supervised control of locomotion mode transition hinges on constant operator-robot interaction, which might not always be feasible or reliable, especially in confined and complex environments typical in search and rescue missions \cite{delmerico2019current}. In such situations, operators might struggle to maintain absolute situational awareness. To address the locomotion mode transition conundrum, various solutions have been proposed. These include adopting specialized mechanical designs \cite{bovzic2017optimization, cao2022omniwheg} and applying pre-programmed solutions \cite{chen2017turboquad}. Although these methods have enhanced the autonomy of locomotion mode transitions, universally applicable autonomous solutions remain in the early stages of development. In fact, most locomotion mode transitions in hybrid robots are currently achieved via high-level human operator control. This applies to cutting-edge wheel/track-legged robots, including DRC-HUBO, CHIMP, Momaro, and RoboSimian, depicted in Fig. \ref{figure:1-1}, which were four of the top five robot designs crafted for the DARPA Robotics Challenge \cite{krotkov2018darpa}.

A major obstacle in achieving seamless autonomous locomotion transition lies in the need for an efficient sensing methodology that can promptly and reliably evaluate the interaction between the robot and the terrain, referred to as terramechanics. These methods generally involve performing comprehensive on-site measurements of soil attributes prior to robot deployment \cite{thomson2012kinematic}. Moreover, it's important to consider that these terramechanics models, striving to predict robot-terrain interactions, often involve substantial computational costs due to their complexity \cite{wong2009terramechanics}. Therefore, terramechanics methods are unsuitable for use in autonomous locomotion mode transition control directly, particularly in scenarios where robots need to move at high speeds, for example in search and rescue missions. To bypass the limitations of terramechanics methods, researchers have probed into alternative strategies for accomplishing autonomous locomotion transition. For example, certain studies have utilized energy consumption as a metric for evaluating the transverse-ability of different locomotion modes in wheel/track-legged robots \cite{leppanen2008sensing}. By scrutinizing the energy expenditure for different locomotion modes, researchers can evaluate their efficiency in navigating various terrains. Additionally, other general parameters like stability margin and motion efficiency have been examined in the quest to achieve autonomous locomotion transition \cite{kobayashi2012locomotion}.

In the literature review, Gorilla \cite{kobayashi2012locomotion} is able to switch between bipedal and quadrupedal walking locomotion modes autonomously using criteria developed based on motion efficiency and stability margin. WorkPartner \cite{leppanen2008sensing} demonstrated its capability to seamlessly transition between two locomotion modes: rolling and rolking. The rolking mode, a combination of rolling and walking, empowered WorkPartner to navigate with enhanced agility. This feat was accomplished through the implementation of devised criteria that took into account a comprehensive analysis of energy utilization, wheel slip percentage, and the intricate dynamics between the wheels and the demanding terrain. However, it's noteworthy that Gorilla only has walking locomotion mode and does not fit into the wheel/track-legged hybrid robot category. It is important to note that the approach introduced by WorkPartner is tailored specifically to it. The threshold values for locomotion transition criteria were established empirically through prior experimental evaluations conducted on the target terrains. However, a critical aspect that deserves emphasis is that the prevailing criteria proposed for locomotion mode transitions have primarily concentrated on the robot's internal states, neglecting the integration of external environmental information into the decision-making process. This oversight underscores the need for future developments that incorporate a more comprehensive understanding of the external context and environmental factors, enabling robots like WorkPartner to make informed decisions based on a holistic assessment of both internal and external conditions.

This paper presents a novel methodology for achieving autonomous locomotion mode transitions in quadruped wheel/track-legged hybrid robots, taking into account both internal states of the robot and external environmental conditions. Our emphasis is on the ``articulated wheel/track robot'' \cite{chen2017turboquad}, where the wheels or tracks are affixed to the joints or leg ends for synchronized movements. A notable example is the state-of-the-art wheel-legged robot, ANYmal with Wheels \cite{bjelonic2021whole}. For such quadruped wheel/track-legged robots negotiating steps, energy efficiency and consumption are critical considerations for devising autonomous mobility strategies \cite{dalvand2006stair, turker2012step}.

The cornerstone of our transition criterion combines energy consumption data with the geometric heights of the steps encountered. These threshold values are determined in energy evaluations while the robot operates in the walking locomotion mode. To analyze the energy dynamics during step negotiation in this mode, we propose two climbing gaits to achieve appropriate locomotion behaviors \cite{wang2016stair}. A distinguishing feature of our approach is its independence from specific mechanical designs \cite{cao2022omniwheg, chen2017turboquad}, rendering it adaptable to a wide array of hybrid robots. Ultimately, our method marks a pivotal advancement in the realm of autonomous mode transitions during step negotiation, as it holistically integrates both internal and external determinants to finalize transition thresholds.

The paper's organization is as follows. The second section presents the two main locomotion methods employed by the Cricket robot, rolling and walking, along with a description of two gaits designed for negotiating steps. In the third section, we outline the mathematical framework used for quantifying energy expenditure. Section four illustrates our novel approach for automating locomotion mode transitions and supports our findings with simulation outcomes. Lastly, section five wraps up the paper with a conclusion and delineates potential avenues for future research.

\section{Locomotion Modes of the Cricket Robot}
\label{sec:locomotion}

\begin{figure}
\vspace{0.3cm}
    \centerline{\includegraphics[width=0.98\columnwidth]{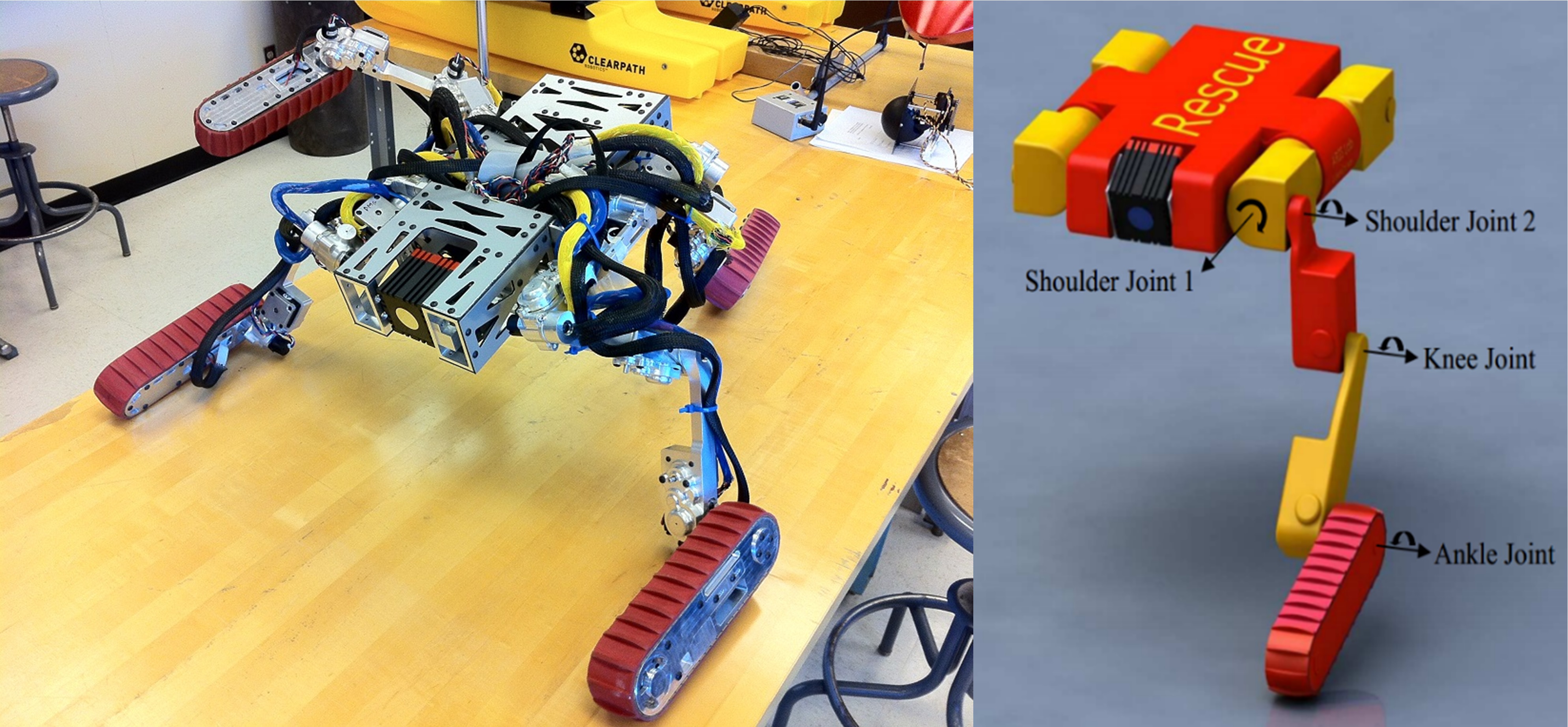}}
    \caption{The Cricket robot (left) and its leg joints layout (right). The Cricket robot \cite{davies2015reconfigurable} is a hybrid locomotion system that utilizes four revolute joints on each leg. The outermost leg segment is equipped with a drivable track that encircles it, enabling the robot to move like traditional skid-steer tank robots.
    }
    \label{figure:1}
\end{figure}

This section describes the primary locomotion modes, rolling and walking locomotion of our hybrid track-legged robot named Cricket shown in Fig. \ref{figure:1}. It also introduces two proposed gaits designed specifically for step negotiation in quadrupedal wheel/track-legged robots.  

\subsection{The Cricket Robot}

\begin{figure}
    \vspace{0.3cm}
    \centerline{\includegraphics[width=0.85\columnwidth]{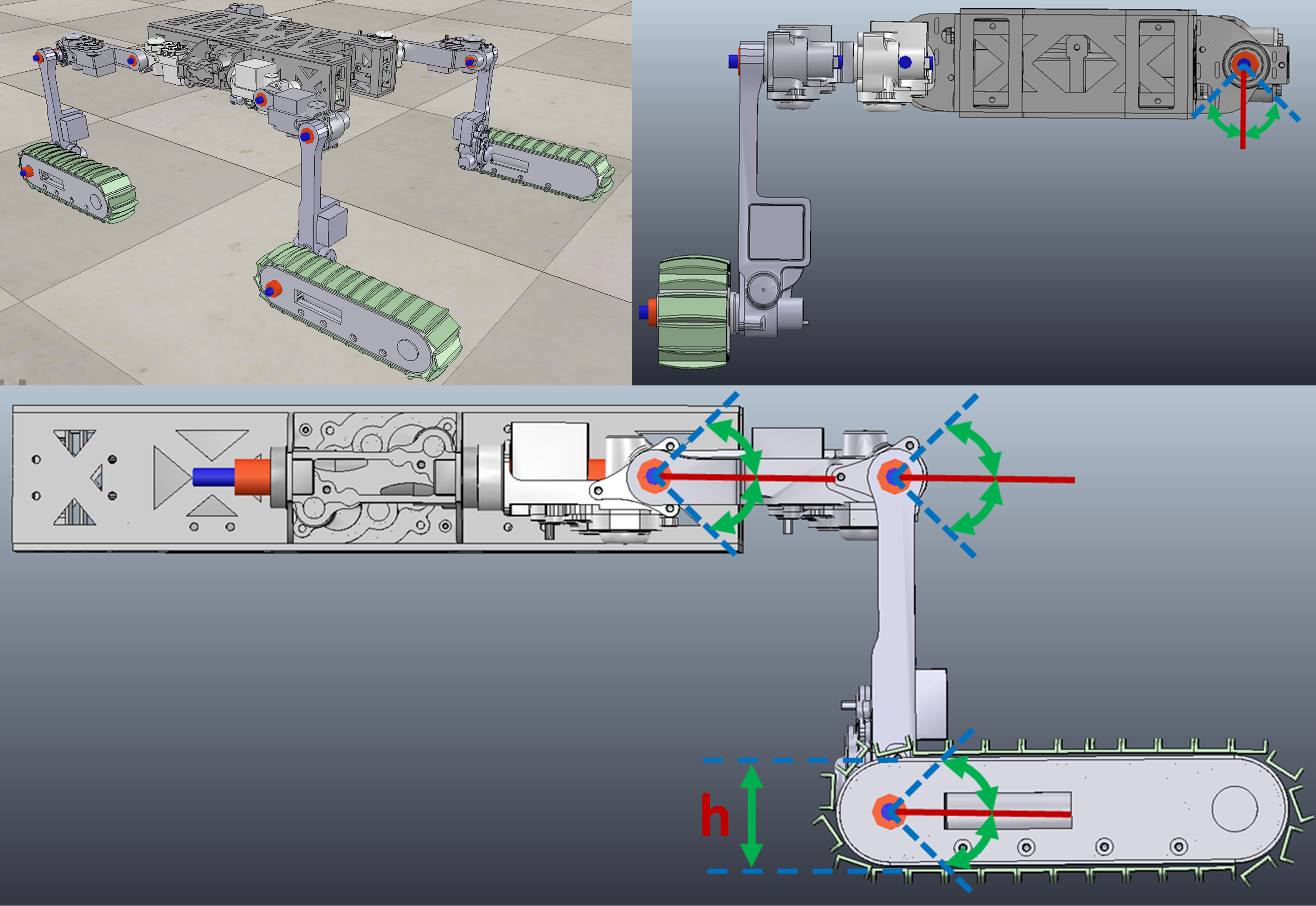}}
    \caption{The configuration and positions of joints in Cricket's legs during rolling locomotion \cite{wang2017autonomous}.  
    }
    \label{figure:2}
\end{figure}

The Cricket robot, as referenced in \cite{davies2015reconfigurable}, forms the basis of this study, being a fully autonomous track-legged quadruped robot. Its design specificity lies in embodying fully autonomous behaviors, and its locomotion system showcases a unique combination of four rotational joints in each leg, which can be seen in Fig. \ref{figure:2}. Moreover, every leg is equipped with a drivable track that circumnavigates the outermost leg segment. This design enables the robot to steer in a manner reminiscent of traditional tank robots. However, unlike its contemporaries, the Cricket robot possesses the ability to conduct intricate movements, such as navigating through uneven terrain, in its walking locomotion mode \cite{davies2011novel}. The two primary forms of the robot's movement are rolling, which leverages tracks for efficient movement across semi-flat terrains, and walking, which is primarily used for maneuvering across challenging and uneven terrains. In this paper, these modes will be referred to as rolling and walking, respectively. Similar to many other hybrid robots, the default locomotion mode of the Cricket robot is rolling. This mode is preferred on flat and rigid surfaces due to its efficiency in terms of time and energy consumption. In the rolling locomotion mode, the robot maintains its home configuration, where all joints are positioned at their central positions as illustrated in Fig. \ref{figure:2}.

\subsection{Walking Locomotion Mode of the Cricket}
\label{sec:walking_gaits}
\begin{figure}
    \vspace{0.3cm}
    \centerline{\includegraphics[width=0.98\columnwidth]{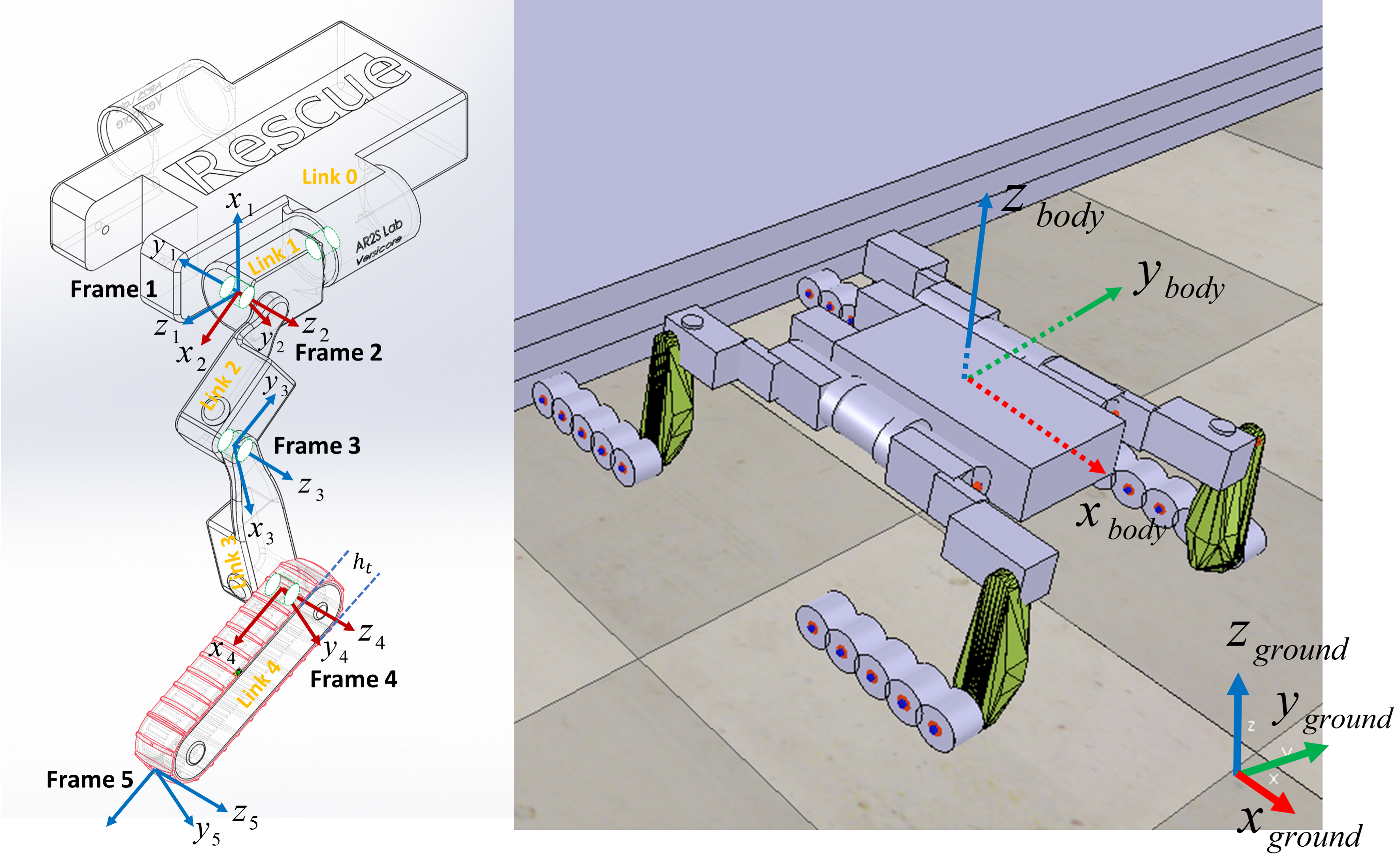}}
    \caption{The assignment of frames for the links of the front left leg of the Cricket robot (left) and the corresponding model created within the CoppeliaSim (formerly V-REP) robotics simulation software (right) \cite{wang2017autonomous}. 
    }
    \label{figure:3}
\end{figure}

In order to successfully traverse steps, the Cricket robot employs specifically designed climbing gaits to ensure smooth walking performance. The process of designing these walking gaits is a combination of adjusting body positions and controlling leg movements, which necessitates the use of inverse kinematics calculations. To establish the kinematic model, we employed the well-known Denavit-Hartenberg (D-H) method \cite{saha2014introduction}. This involved setting up coordinate frames for each link of the legs and the robot's body, as illustrated in Fig. \ref{figure:3}.
Table \ref{tab:simulation_metrics} details the D-H parameters pertaining to the robot's left leg in the front. It is worth noting that the remaining three legs of the Cricket robot share similar parameters. This symmetry is a result of the mechanical design of the robot, allowing for consistent and uniform characteristics among these legs.
\begin{table}
    \caption{The D-H parameters specific to the left leg in the front of the Cricket robot }
    \label{tab:simulation_metrics}
    \begin{center}
        \begin{tabular}{|c|c|c|c|c|}
           \hline Link & $\boldsymbol{b}_{\boldsymbol{i}}$ (m) & $\boldsymbol{\theta}_{\boldsymbol{i}}$ (deg.) & $\boldsymbol{a}_{\boldsymbol{i}}$ (m) & $\boldsymbol{\alpha}_{\boldsymbol{i}}$ (deg.) \\
           \hline 1 & 0 & $\theta_1$ & 0 & $\pi / 2$ \\
            \hline 2 & $b_2=0.1020$ & $\theta_2$ & $a_2=0.1330$ & 0 \\
            \hline 3 & $b_3=0.0185$ & $\theta_3$ & $a_3=0.1850$ & 0 \\
            \hline 4 & $b_4=0.0285$ & $\theta_4$ & $a_4=0.2196$ & 0 \\
            \hline
        \end{tabular}
    \end{center}
\end{table}
The homogeneous transformation matrix connecting the shoulder frame (Frame 1) to the track tip (Frame 5) as shown in Fig. \ref{figure:3} was computed as
\begin{ceqn}
    \begin{align}
        { }^1 T_5 ={ }^1 T_2{ }^2 T_3{ }^3 T_4{ }^4 T_5=\prod_1^5{ }^i T_{i+1} = \qquad \qquad \qquad \qquad \quad \nonumber
    \end{align}
\end{ceqn}
\vspace{-0.5cm}
\begin{multline}
 \left[
  \begin{matrix}
        c_1 c_{234} & -c_1 s_{234} & s_1 & \left(b_2+b_3+b_4\right) s_1+c_1\left(a_2 c_2+a_3 c_{23}+a_4 c_{234}\right) \\
        s_1 c_{234} & -s_1 s_{234} & -c_1 & -\left(b_2+b_3+b_4\right) c_1+s_1\left(a_2 c_2+a_3 c_{23}+a_4 c_{234}\right) \\
        s_{123} & c_{123} & 0 & a_2 s_2+a_3 s_{23}+a_4 s_{234} \\
        0 & 0 & 0 & 1 
  \end{matrix}\right] , \\
  \tag{1} \label{eqn:d-h} 
\end{multline}
where $s_i$ and $c_i$ represent $\sin \theta_i$ and $\cos \theta_i$ respectively, $s_{i j}$ and $c_{i j}$ represent $\sin \left(\theta_i+\theta_j\right)$ and $\cos \left(\theta_i+\theta_j\right)$ respectively, and $s_{i j k}$ and $c_{i j k}$ represent $\sin \left(\theta_i+\theta_j+\theta_k\right)$ and $\cos \left(\theta_i+\right.$ $\left.\theta_j+\theta_k\right)$ respectively.

\begin{figure}
    \vspace{0.3cm}
    \centerline{\includegraphics[width=0.98\columnwidth]{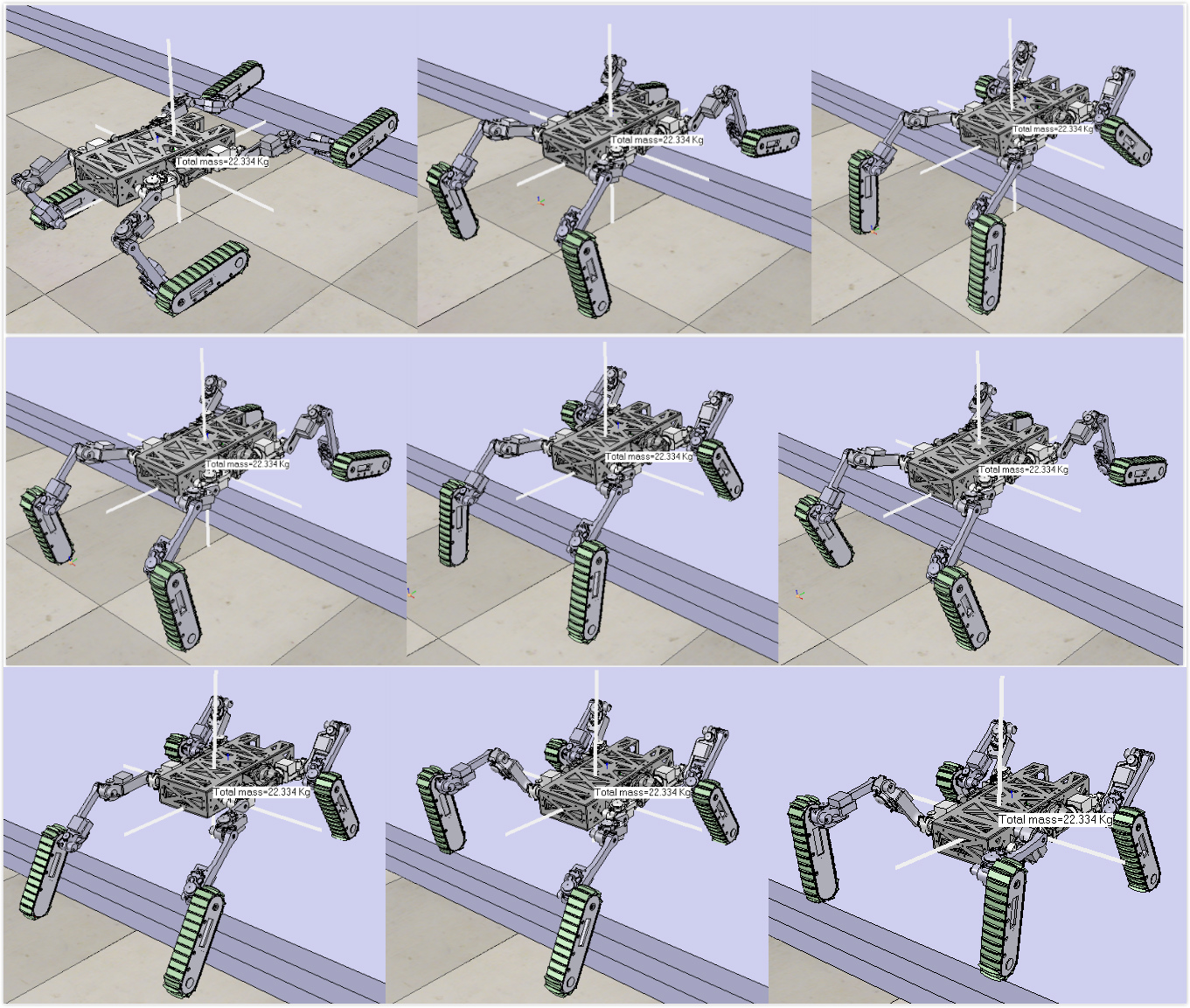}}
    \caption{Visualization of the whole body climbing gait for step negotiation in CoppeliaSim \cite{wang2016stair}.  
    }
    \label{figure:4}
\end{figure}

\begin{figure}
    \vspace{0.3cm}
    \centerline{\includegraphics[width=0.7\columnwidth]{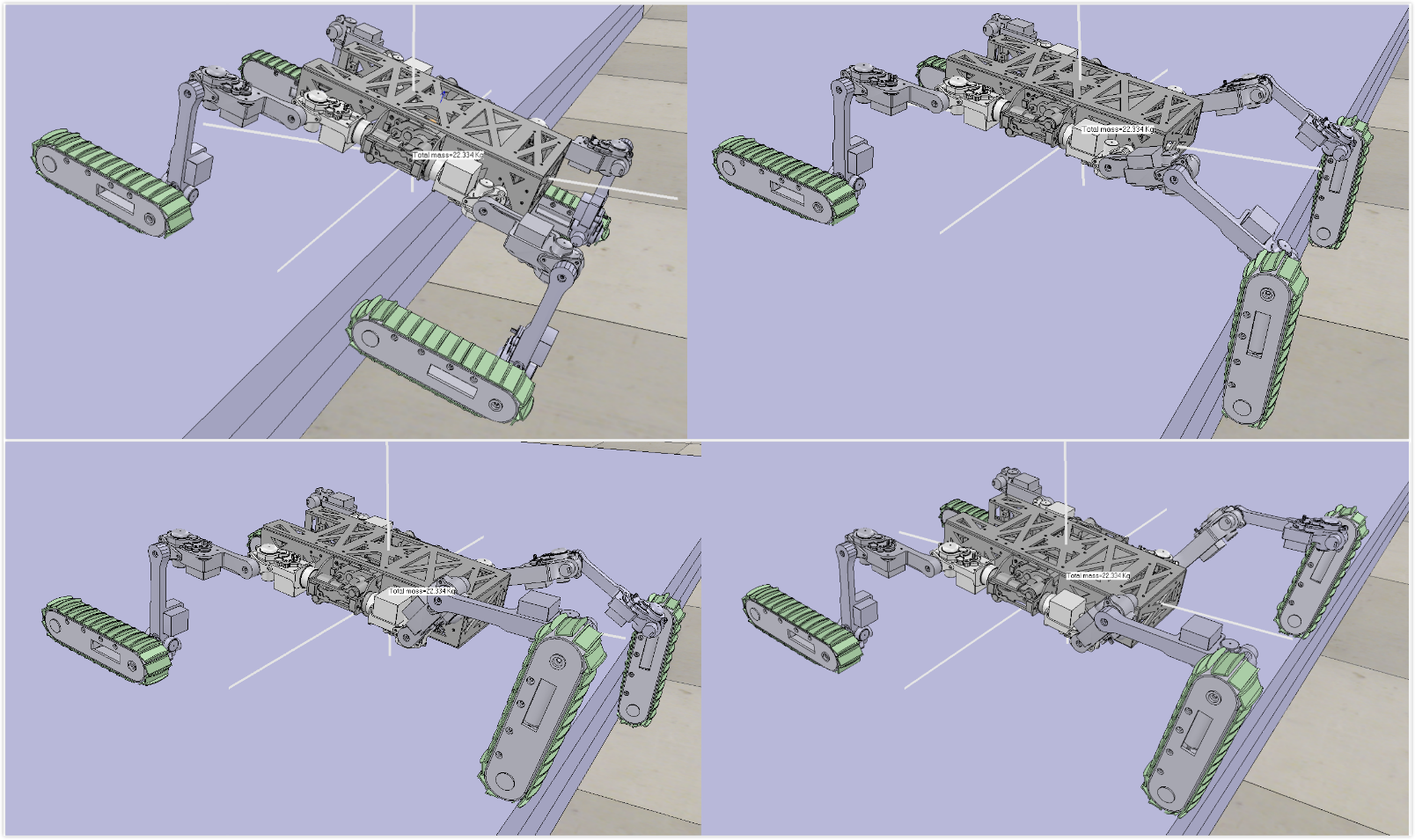}}
    \caption{Visualization of the rear body climbing gait for step negotiation in CoppeliaSim \cite{wang2016stair}.  
    }
    \label{figure:5}
\end{figure}

While the study of legged locomotion gaits has been a topic of research for several decades, the investigation of locomotion in wheel-legged robots is a relatively recent area of study \cite{thomson2012kinematic}. Hybrid ground robots, equipped with highly articulated legs with more than three degrees-of-freedom, present unique challenges in gait development. Our study contributes to this growing field by suggesting two novel climbing gaits to surmount steps of different dimensions (h, 2h, and 3h, where h represents the track height as displayed in Fig. \ref{figure:2}). We term these the whole-body climbing gait and the rear-body climbing gait \cite{wang2016stair}, demonstrated in Fig. \ref{figure:4} and Fig. \ref{figure:5}, respectively.

The track tip positioning was the key parameter controlled during the creation of these climbing gaits. To assure seamless locomotion, trajectories for each joint of the robot were defined through a fifth-order polynomial along with their first and second derivatives. The trajectory design took into account six constraints: initial and final position, velocity, and acceleration \cite{roy2014kinematics}. The Reflexxes Motion Library IV \cite{kroger2011opening} was utilized to perform the inverse kinematics calculation.

The whole-body climbing gait involves utilizing the entire body movement of the robot, swaying forwards and backwards to enlarge the stability margins before initiating gradual leg movement to overcome a step. This technique optimizes stability during the climbing process. To complement this, the rear-body climbing gait was developed. In this approach, once the front legs and body have completed their upward rolling motion, the rear legs are elevated to ascend the step. This strategy is particularly beneficial in situations where the mobility of rolling locomotion is hindered by the rear wheels. For a more detailed discussion of the whole-body climbing gait and the rear-body climbing gait, we direct readers to \cite{wang2016stair}.

\section{Computation of Energy Usage}
\label{sec:energy}

In order to account for the robot's dynamics and precisely quantify energy consumption during step negotiation, we utilized the Vortex physical engine incorporated within CoppeliaSim (previously known as V-REP) robotics simulation software \cite{rohmer2013v}. This ensured robust management of the robot's intricate dynamics and interactions. Our choice of CoppeliaSim as the robot modeling and simulation environment is grounded in its consistent record of high-fidelity simulations. A recent comparative analysis of four leading mobile robotics simulators---CoppeliaSim, Gazebo, MORSE, and Webots---underscored CoppeliaSim's unmatched accuracy in motion simulation. Moreover, CoppeliaSim is distinguished by its adaptability, accommodating an extensive variety of programming languages, physics engines, and model types \cite{farley2022pick}.

For the quantification of energy used by a DC motor across a specific time duration $T$, we used the following formula \cite{roy2014kinematics}:
\begin{ceqn}
    \begin{equation}
        E=\int_0^T U_a I_a d t=\int_0^T \tau \dot{\theta} d t+\int_0^T I_a^2 R_a d t \, , \tag{2} \label{eqn:energy_gen} 
    \end{equation}
\end{ceqn}
In this equation, $U_a$ symbolizes the applied voltage, and $I_a$ stands for the armature current. $\tau$ denotes joint torque and $\dot{\theta}$ is the joint angular velocity. $R_a$ signifies the armature resistance. $\tau$ represents the joint torque that is calculated using $\tau = K_t I_a$, with $K_t$ as the torque constant. Equation \eqref{eqn:energy_gen} helps compute the total energy expenditure of the DC motor within time duration $T$, accounting for both mechanical energy and energy dissipation through heat. When the first term, representing mechanical energy, has a negative value, it suggests that an external force is indeed supplying net energy. However, as discussed in \cite{roy2014kinematics}, this energy can not be stored in DC motors. Hence, the calculation for the total energy used by the DC motor during a specified time duration $T$ is as follows:
\begin{ceqn}
    \begin{equation}
        E=\int_0^T[f(\tau \dot{\theta})] d t+\int_0^T I_a^2 R d t \, , \tag{3} \label{eqn:energy} 
    \end{equation}
\end{ceqn}
where $f(\tau \dot{\theta})=\left\{\begin{array}{cl}\tau \dot{\theta} & \text { when } \tau \dot{\theta}>0 \\ 0 & \text { when } \tau \dot{\theta} \leq 0\end{array}\right.$.
For the calculation of energy expenditure during step negotiation, equations \eqref{eqn:energy_gen} and \eqref{eqn:energy} are combined:
\begin{ceqn}
    \begin{equation}
        E_{\text {total }}=\sum_{i=1}^n E_i=\int_0^T \sum_{i=1}^n\left(f\left(\tau_i \dot{\theta}_i\right)+\frac{\tau_i^2}{K_t^2} R_a\right) d t \, , \tag{4} \label{eqn:energy_total} 
    \end{equation}
\end{ceqn}
In this equation, $T$ is the time taken to negotiate, $E_i$ is the energy used by the actuated joint $i$, $n$ denotes the actuated joint number, and $dt$ signifies the time step.

\section{Autonomous Locomotion Mode Transition}
\label{sec:transition}
In this section, we explore the autonomous locomotion mode transition of the Cricket robot. We present our hierarchical control design, which is simulated in a hybrid environment comprising MATLAB and CoppeliaSim. This design facilitates the decision-making process when transitioning between the robot's rolling and walking locomotion modes. Through energy consumption analyses during step negotiations of varied heights, we establish energy criterion thresholds that guide the robot's transition from rolling to walking mode. Our simulation studies reveal that the Cricket robot can autonomously switch to the most suitable locomotion mode based on the height of the steps encountered. 

It is important to emphasize that the locomotion mode transitions are only meaningful when both rolling and walking modes are capable of handling a step negotiation. And in the step negotiation simulations, it has been observed that the rolling locomotion can not transverse over steps with height more than three time of the track height due to the track traction forces limitation \cite{wang2017autonomous}. As such, both our energy consumption evaluations and autonomous mode transition simulations have been confined to step heights manageable by both locomotion modes. 

\subsection{Simulation Settings}

\begin{figure}
    \vspace{0.3cm}
    \centerline{\includegraphics[width=0.96\columnwidth]{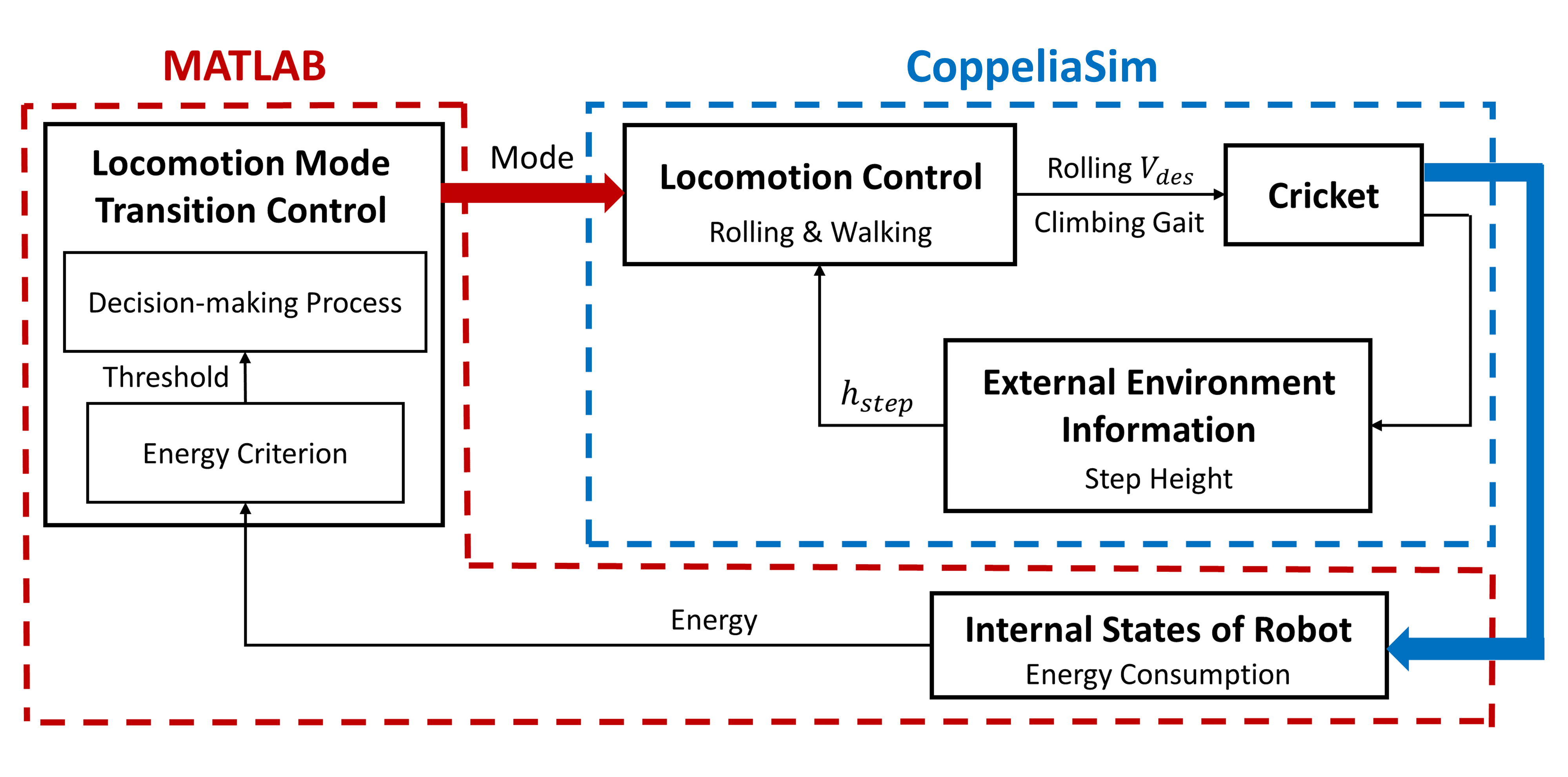}}
    \caption{Hierarchical control design for autonomous locomotion mode transition \cite{wang2017autonomous}.
    }
    \label{figure:7}
\end{figure}

Fig. \ref{figure:7} illustrates the hierarchical control design for the autonomous locomotion mode transition. The decision-making process for this transition is accomplished in MATLAB, whereas the control of each separate locomotion mode is enacted in CoppeliaSim. The connection between MATLAB and the physical robot model in CoppeliaSim is facilitated through the use of the remote API function available in the CoppeliaSim environment. Within CoppeliaSim, control is applied to rolling locomotion in order to maintain the required vehicle speed and home configuration. As for walking locomotion, the climbing gaits created from the step height data, as discussed in Sec. \ref{sec:walking_gaits}, are employed. In order to facilitate motion control in both locomotion modes, all the necessary kinematics and dynamics calculations are carried out within the CoppeliaSim simulation environment. This includes computing torques and angular velocities for each joint. The simulation outputs, along with these calculated values, are then sent back to MATLAB for further data analysis and energy usage calculations. During the step negotiation simulations, a timestep of 2 milliseconds is employed to simulate real-time dynamics accurately.

\subsection{Walking Locomotion Energy Evaluation}

\begin{figure}
    \vspace{0.3cm}
    \centerline{\includegraphics[width=0.98\columnwidth]{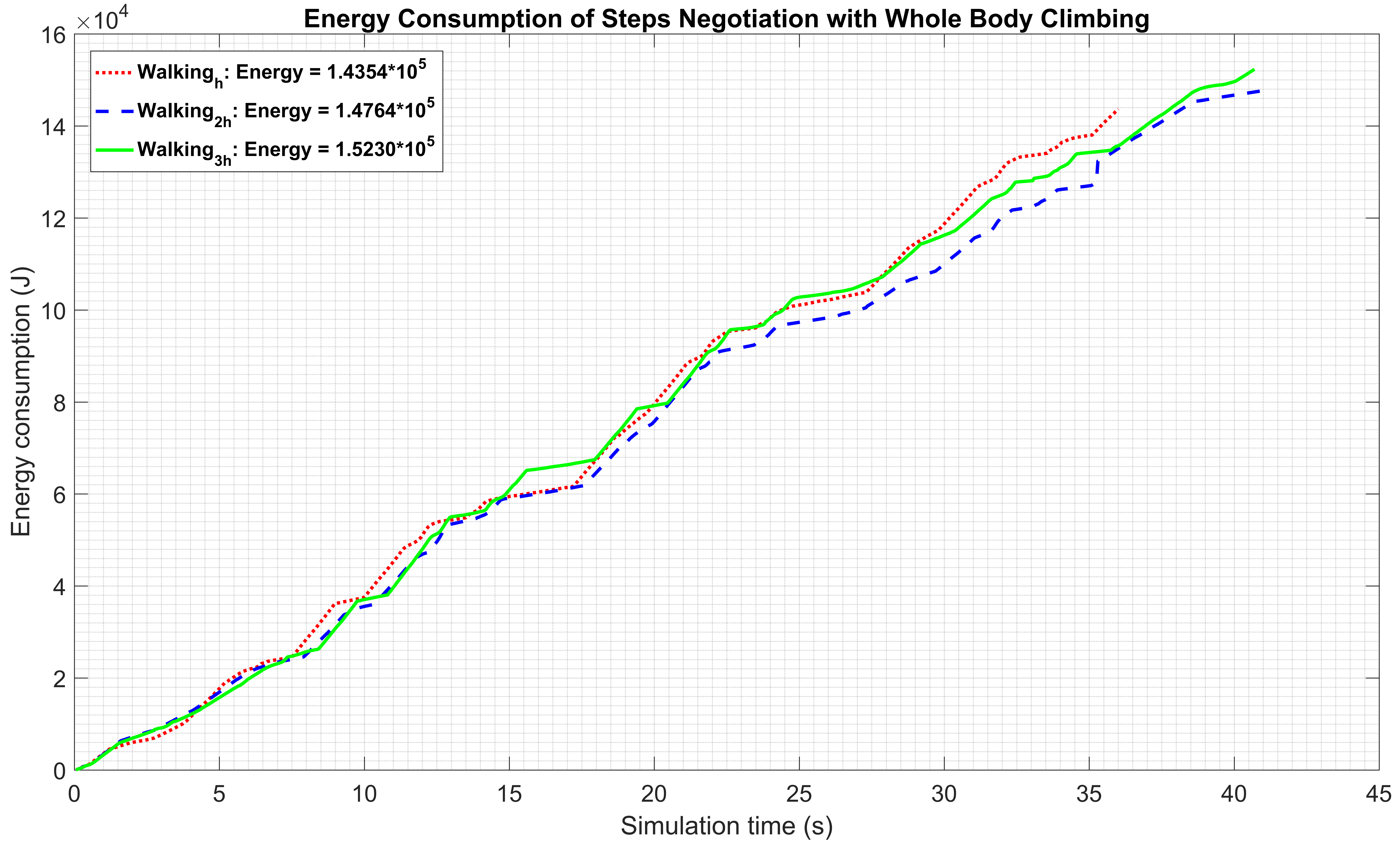}}
    \caption{Total energy consumed while using the whole body climbing gait to negotiate steps of heights h, 2h, and 3h. These values inform the threshold values of the energy criterion for locomotion mode transition control \cite{wang2017autonomous}.
    }
    \label{figure:9}
\end{figure}

\begin{figure}
    \centerline{\includegraphics[width=0.98\columnwidth]{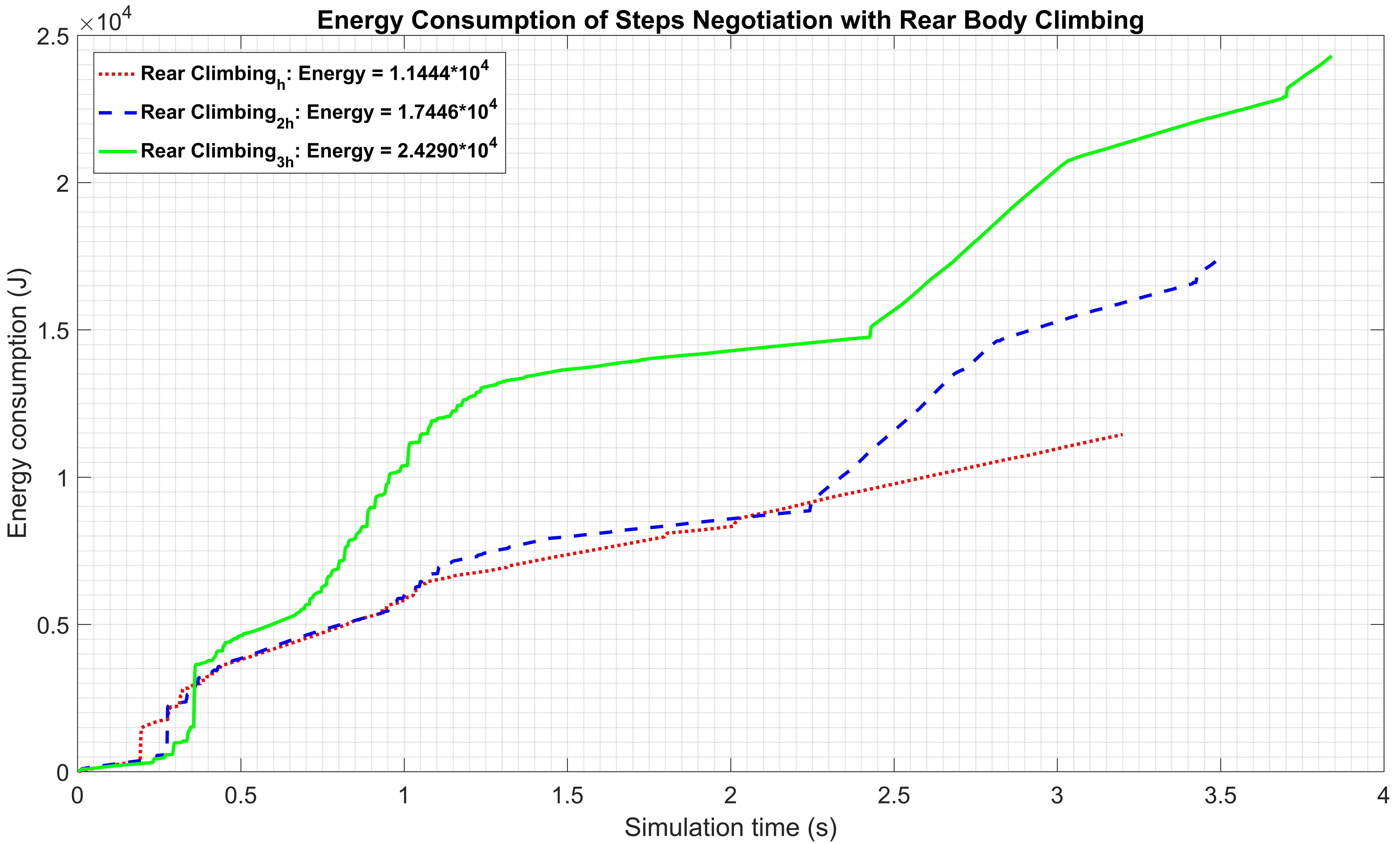}}
    \caption{Accumulated energy consumption while negotiating steps using the rear body climbing gait at heights h, 2h, and 3h.
    }
    \label{figure:10}
\end{figure}

During the step negotiation simulations, it was noticed that the rolling locomotion mode encountered constraints when attempting to cross steps with a height greater than thrice the track height (h being the track height as shown in Fig. \ref{figure:2}). This limitation originates from the traction forces generated by the tracks. As a result, successful locomotion mode transitions can only occur when both rolling and climbing locomotion modes are capable of handling a step negotiation task. For evaluating the energy expenditure during step negotiation, energy assessments were carried out for step heights of h, 2h, and 3h using both the whole body climbing and rear body climbing gaits and shown in Fig. \ref{figure:9} and Fig. \ref{figure:10}, respectively. 

The evaluation of energy consumption for the walking locomotion mode encompassed the entire step negotiation process, from the commencement of the negotiation until its completion. Fig. \ref{figure:9} reveals minimal discrepancies in energy consumption for the whole-body climbing gait, which can be attributed to the thoughtful design of the climbing gaits. These gaits incorporate identical desired joint accelerations, leg stride length, and forward movement height, as highlighted in \cite{wang2017autonomous}. Consequently, variations in energy consumption during different step negotiations primarily stem from negotiation time and body movements. In order to establish the threshold values ($T_{wb}$ and $T_{rb}$) for the energy criterion, they were equated to the energy expenditure of the walking locomotion mode, utilizing the whole-body climbing and rear-body climbing gaits, respectively. To identify the threshold values ($T_{wb}$ and $T_{rb}$) for the energy criterion, they were set equal to the energy expenditure of the walking locomotion mode using the whole body climbing and rear body climbing gaits, respectively. Unlike other methods that use empirical values \cite{kobayashi2012locomotion, leppanen2008sensing}, the threshold values in this study were decided upon based on a novel rule that evaluates the alternative locomotion mode. Moreover, these threshold values are not fixed and are determined based on the terrain profiles the robot is negotiating.

\subsection{Simulation Results}

\begin{figure}
    \vspace{0.21cm}
    \centerline{\includegraphics[width=0.98\columnwidth]{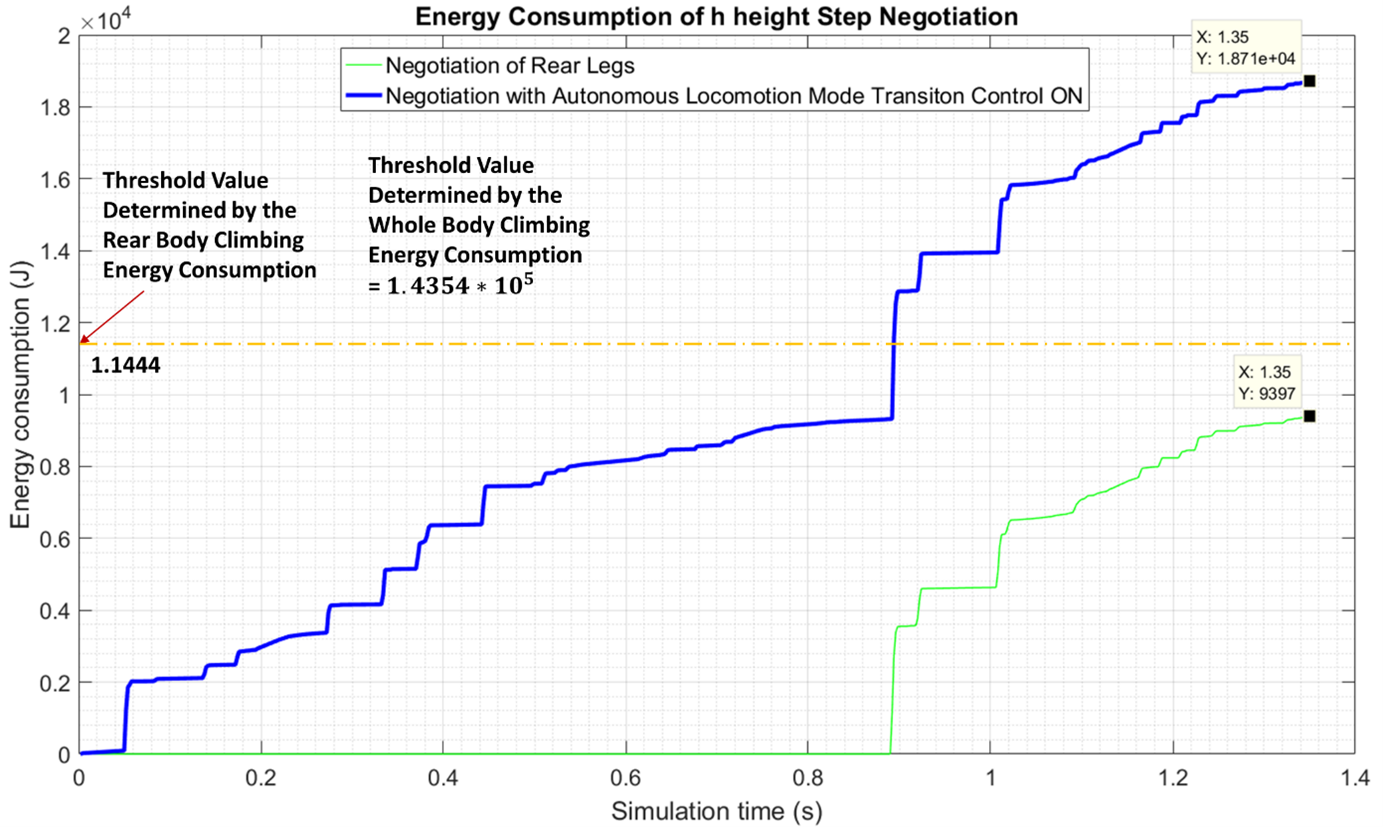}}
    \caption{The Cricket robot tackles a step of height h using rolling locomotion mode, negating the need for a transition to the walking mode. The total energy consumed throughout the entire step negotiation process in rolling locomotion stayed below the preset threshold value. This threshold value was established based on the whole body climbing gait at height h, as shown in Fig. \ref{figure:9}, or the rear body climbing gait at height h, as seen in Fig. \ref{figure:10}. The blue line illustrates the total energy consumed (in rolling locomotion mode), while the green line represents the ongoing cumulative energy consumption of the rear legs, indicating it did not exceed the threshold values set by the rear body climbing gait.
    }
    \label{figure:11}
\end{figure}

\begin{figure}
    \vspace{0.21cm}
    \centerline{\includegraphics[width=0.98\columnwidth]{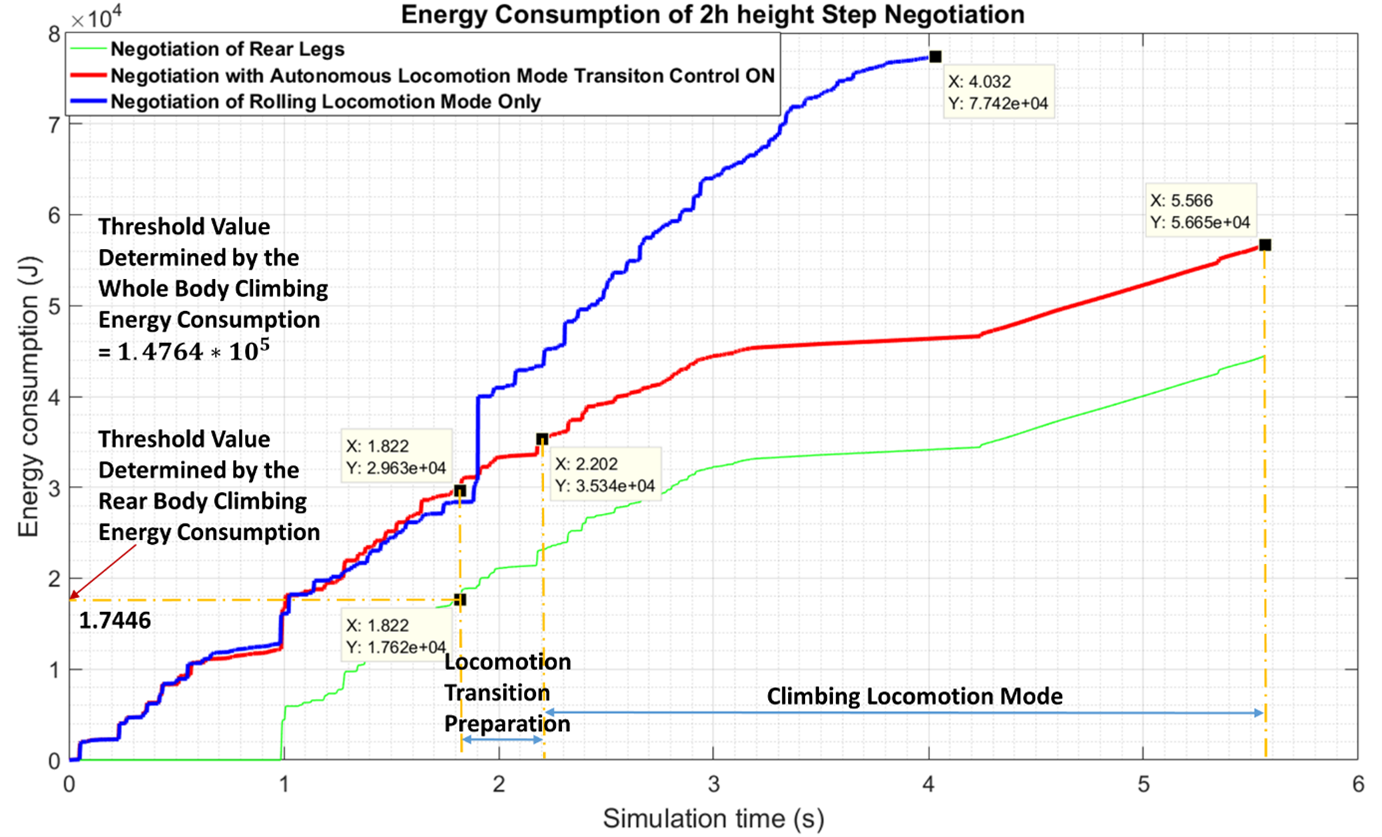}}
    \caption{The Cricket robot tackles a step of height 2h, beginning in rolling locomotion mode and transitioning to walking locomotion mode using the rear body climbing gait. The red line in the plot shows that the robot tackled the step in rolling locomotion mode until the online accumulated energy consumption of the rear legs (depicted by the green line) exceeded the predetermined threshold values set by the rear body climbing gait for heights of 2h. The overlap between the red line (ongoing energy consumption of the robot) and the blue line (pre-studied energy consumption of step negotiation in rolling locomotion mode only) illustrates this. After the mode transition is triggered, the robot enters a well-defined preparation phase, wherein it moves backward a short distance to ensure the rear tracks are separated from the step. Following the preparation phase, the robot switches to the rear body climbing gait. Despite the noticeable improvement in energy consumption, the transition to the rear body climbing gait takes more time for the robot to tackle a 2h step.
    }
    \label{figure:12}
\end{figure}

\begin{figure}
    \vspace{0.21cm}
    \centerline{\includegraphics[width=0.98\columnwidth]{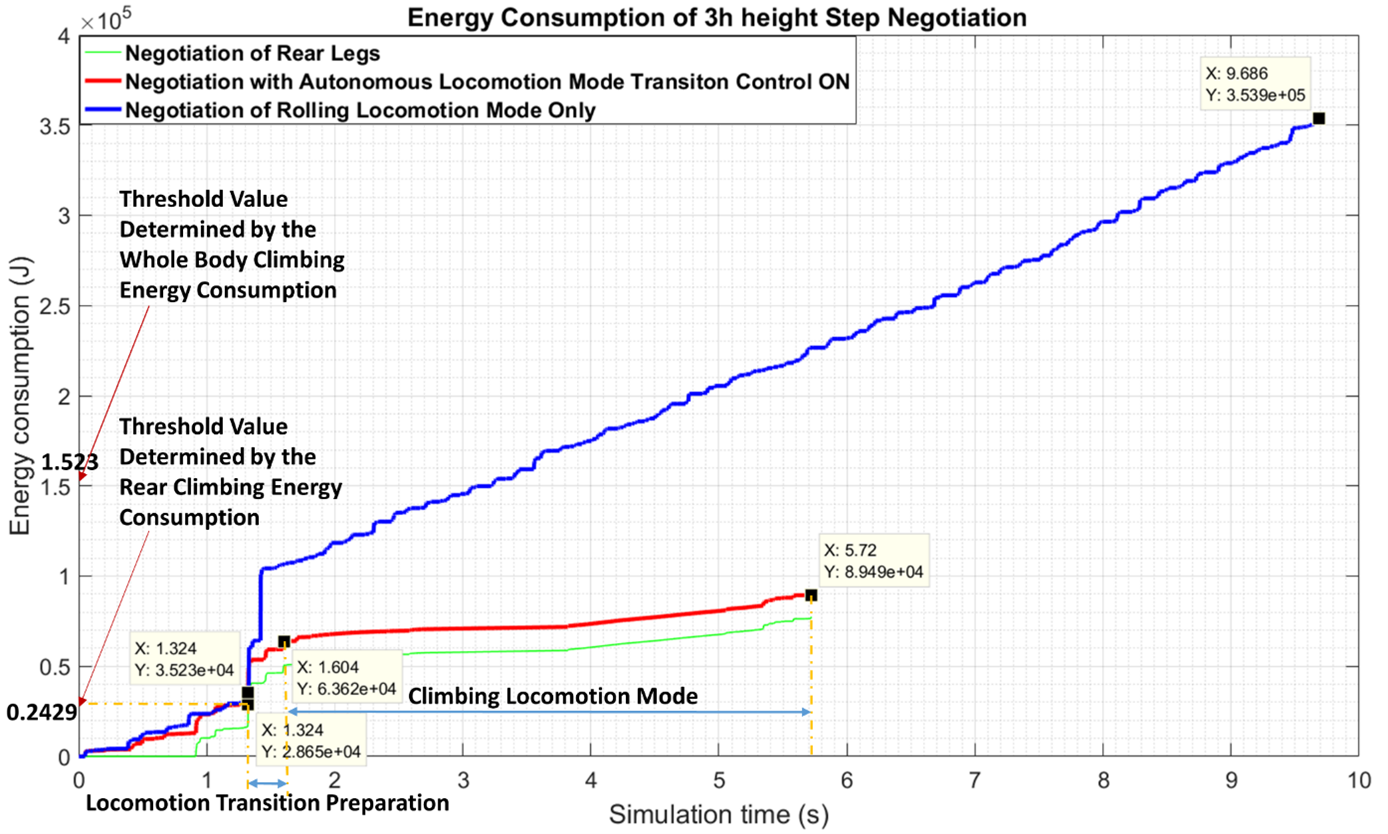}}
    \caption{The Cricket robot tackles a step of height 3h by initiating in rolling locomotion mode and transitioning to walking locomotion mode using the rear body climbing gait. The transition process mirrors that of the 2h step negotiation shown in Fig. \ref{figure:12}. Unlike tackling a 2h step, the robot achieves considerable improvements in both energy consumption and time spent when tackling a step with a height of 3h.
    }
    \label{figure:13}
\end{figure}

To assess the efficacy of the suggested autonomous locomotion mode transition strategy, simulation experiments featuring step heights of h, 2h, and 3h were conducted. These simulations involved continuous tracking of energy consumption for both total body negotiation ($E_{Rw}$) and rear track negotiation ($E_{Rr}$). These values were then compared to the pre-assessed energy results for full body climbing ($E_{Cw}$) and rear body climbing ($E_{Cr}$), presented in Fig. \ref{figure:9} and Fig. \ref{figure:10}, respectively. In addition, to demonstrate the energy advantages of the proposed approach, the energy consumption of step negotiation solely in the rolling locomotion mode (without implementing the proposed method) Fig. \ref{figure:11}, \ref{figure:12}, and \ref{figure:13} illustrate these comparisons for step heights of h, 2h, and 3h correspondingly. These figures provide insight into the patterns of energy consumption, accentuating the efficiency of the proposed strategy in enabling energy-conscious step negotiation. The complete process of autonomous locomotion mode transition during the 2h step negotiation is demonstrated in the following video: \url{https://youtu.be/PGgzRK59ZZI}.


As depicted in Fig. \ref{figure:11}, for the step negotiation operation with a height of $h$, both $E_{R w}<E_{C w}$ and $E_{R r}<E_{C r}$ conditions were met. As a result, the robot successfully completed the step negotiation task using the rolling locomotion mode alone, eliminating the necessity to switch to another locomotion mode. Fig. \ref{figure:12} shows that while the robot was tackling a 2h step, the suggested locomotion mode transition was triggered when the energy consumption of the rear track negotiation $\left(E_{R r}\right)$ surpassed the previously analyzed energy evaluation outcomes of the rear body climbing gait $\left(E_{C r}\right)$ shown in Fig. \ref{figure:10}. To transition from the rolling to walking locomotion mode, a carefully orchestrated preparation phase was employed. During this phase, the robot executed a brief backward movement to create a clear path for the rear track in view of the approaching step, paving the way for a smooth transition from rolling to walking mode.

Similarly, when the robot encountered a step with a height of 3h (as shown in Fig. \ref{figure:13}), the mode transition was activated when the energy consumption of the rear track negotiation in rolling mode surpassed the threshold value derived from the previously assessed energy results of the rear body climbing gait. The results highlight the superior energy efficiency of the proposed autonomous locomotion mode transition method for negotiating steps of different heights, in contrast to solely depending on the rolling locomotion mode. This underscores the efficiency of the proposed strategy in enabling energy-conscious step negotiation across various terrains and obstacles.

\subsection{Summary of simulation results}

The implementation of the energy criterion strategy has proven effective in facilitating autonomous locomotion mode transitions for the Cricket robot when negotiating steps of varying heights. Compared to step negotiation purely in rolling locomotion mode, the proposed strategy demonstrated significant enhancements in energy performance, particularly for taller steps. A significant feature of this method is the determination of transition criterion threshold values based on studies of alternative locomotion modes rather than relying on empirical settings. This contribution is crucial as it ensures a more systematic and objective approach to setting the thresholds for locomotion mode transitions.

\section{Conclusions}
\label{sec:conclusion}

This paper introduced an energy-centric method for automatic transitioning between locomotion modes in quadruped track-legged robots during step negotiation. Exhibiting flexibility, our methodology could be applied to a diverse range of wheel/track-legged robots, deriving transition thresholds from energy assessments during the robot's walking mode. 

Several avenues for future exploration are suggested based on the constraints of the work discussed in this paper. Firstly, conducting more exhaustive energy evaluations of walking locomotion for varying step heights could provide a more nuanced understanding of energy consumption patterns and facilitate the development of an exhaustive look-up table. Furthermore, despite the common usage of climbing gaits for step negotiation in wheel/track-legged robots, it is crucial to validate the energy efficiency of the suggested climbing gaits. Future efforts could concentrate on refining the climbing gait to further minimize energy consumption and enhance overall performance. Lastly, investigating alternative locomotion strategies (such as simultaneous rolling and walking \cite{bjelonic2021whole}) and assessing their energy efficiency may offer crucial insights for forthcoming advancements in the control of locomotion mode transition.




\bibliographystyle{elsarticle-num} 
\bibliography{bibliography}






\end{document}